\documentclass[letterpaper]{article} 
\usepackage{aaai2026}  
\usepackage{times}  
\usepackage{helvet}  
\usepackage{courier}  
\usepackage[hyphens]{url}  
\usepackage{graphicx} 
\usepackage{natbib}  
\usepackage{caption} 
\usepackage{algorithm}
\usepackage{algorithmic}

\usepackage{newfloat}
\usepackage{listings}
\DeclareCaptionStyle{ruled}{labelfont=normalfont,labelsep=colon,strut=off} 
\floatstyle{ruled}
\newfloat{listing}{tb}{lst}{}
\floatname{listing}{Listing}
\usepackage{booktabs}
\usepackage{tabularx}
\usepackage[table]{xcolor}

\usepackage{afterpage}
\usepackage{placeins}

\title{MonitrLLM: A Community-Centered Evaluation Infrastructure for Large Language Models}
\author {
    Victor Ojewale\textsuperscript{\rm 1},
    Ro Encarnación\textsuperscript{\rm 2},
    Suresh Venkatasubramanian\textsuperscript{\rm 1}
    Danaé Metaxa\textsuperscript{\rm 2}
}
\affiliations {
    \textsuperscript{\rm 1}Center for Tech Responsibility, Brown University\\
    \textsuperscript{\rm 2}University of Pennsylvania\\

}

\begin{document}

\maketitle

\begin{abstract}
Benchmark suites assess model capability on controlled tasks; large-scale conversation corpora capture naturalistic use without user feedback; and in-interface feedback mechanisms record satisfaction without task purpose. Together, they leave a critical gap in LLM evaluation: no existing infrastructure routinely links interaction trajectories to user-defined outcomes. We introduce \textbf{MonitrLLM}, open-source infrastructure for community-centered LLM evaluations that links full conversation transcripts to user-reported task intent and outcome assessments, treating all three as primary evaluative signals rather than optional metadata.
To demonstrate the value of this approach, we conducted a two-week feasibility pilot with 26 college students using ChatGPT, collecting 206 evaluation reports with full conversation transcripts. The findings from our pilot demonstrate the value of connecting conversation trajectories with user-reported outcomes. For instance, despite reporting high average satisfaction (4.19/5) with their LLM interactions, participants also experience a substantial 23.1\% failure rate on their goal tasks. We also find that multi-turn conversations are reported as failing at 2.5 times the rate of single-turn exchanges, a pattern that reframes extended interaction as a signal of difficulty rather than engagement. We conclude by discussing the value of incorporating direct user feedback with observational data for robust LLM evaluations, and the possibilities for infrastructure that enables this goal. 
\end{abstract}

\section{Introduction}
Large language models (LLMs) are increasingly being embedded in everyday workflows \citep{nazi_large_2024, li_large_2023, lai_large_2024}. Their widespread use has sharpened concerns about bias \citep{10.1145/3597307}, misinformation \citep{10.1145/3703155}, overconfident answers \citep{10.1145/3706598.3713408}, and uneven performance across user groups \citep{10.1145/3715275.3732212}. These concerns have driven intensive evaluation efforts, including broad benchmark suites that measure capabilities and risks across tasks and metrics \citep{liang2023holisticevaluationlanguagemodels}, and organizational red teaming that probes deployed model behavior \citep{perez2022redteaminglanguagemodels}. Yet many problematic model behaviors and mismatches only become visible when models are used for situated goals, under real constraints, with users adapting their strategies over time \citep{le-jeune-etal-2025-realharm}. 

A central reason these mismatches go undetected is infrastructural: the feedback mechanisms built into deployed LLM interfaces (thumbs up or down reactions, optional free-text fields) record whether users found a response satisfying in the moment, but do not capture what they were trying to accomplish, whether the output was usable for their specific context, or how much interaction effort the exchange required.
Large-scale efforts collect real-world human--LLM conversations via public datasets (e.g., OpenAssistant, WildChat) and data donation tools such as ShareLM, enabling researchers to study interaction patterns across diverse users and deployment contexts \citep{köpf2023openassistantconversationsdemocratizing,zhao2024wildchat1mchatgptinteraction,don-yehiya-etal-2025-sharelm}. While these efforts expand the data available for evaluation, they too usually lack the needs and governance priorities of any particular user community. Conversations are treated as generic data points, and users have limited influence over which evaluation questions are asked, what constitutes success, and how findings are returned to the community.

\paragraph{Our Work.}
In this paper, we argue for \textit{community-centered evaluation} that takes a different starting point. The community's tasks and the community's definitions of success, rather than a generic benchmark target, are the primary evaluative objects. We do this through two contributions.

\begin{figure}[t]
  \centering
  \includegraphics[width=0.50\textwidth, keepaspectratio]{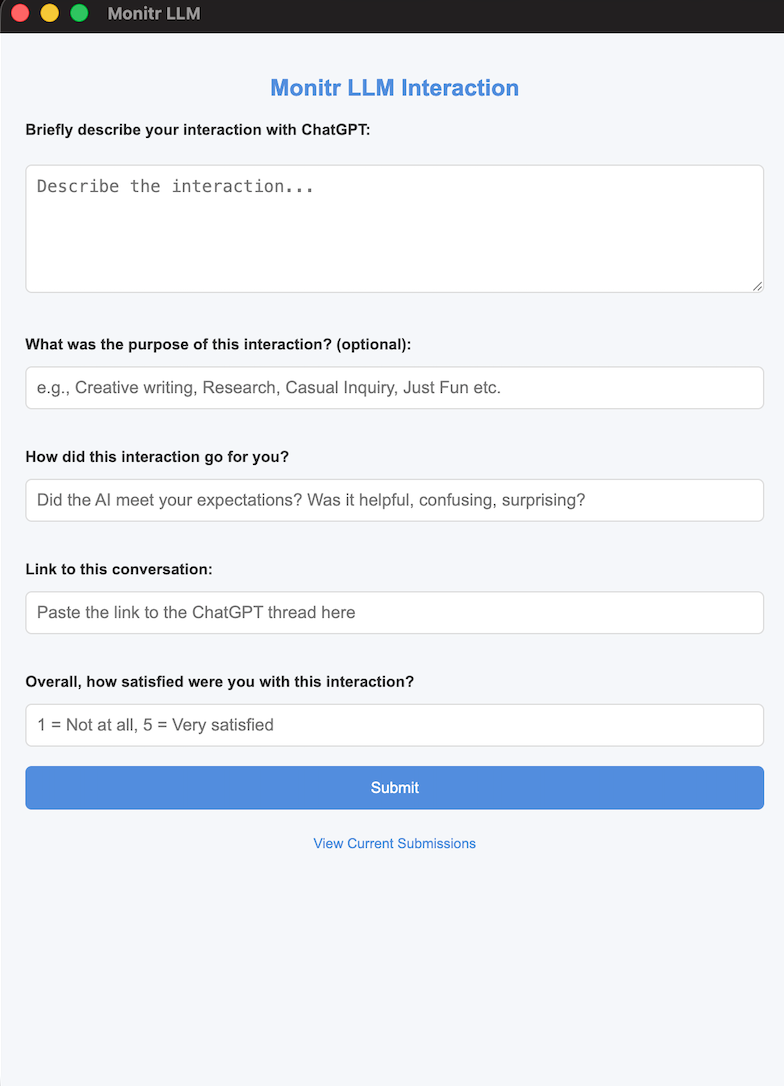}
  \caption{The MonitrLLM browser extension where users submit an audit report through a five-field form capturing interaction description, task purpose, outcome assessment, conversation share link, and a 1-to-5 satisfaction rating.}
  \label{fig:extension}
\end{figure}

First, we introduce \textbf{MonitrLLM}, infrastructure for community-centered LLM evaluations that builds on this principle by linking full conversation transcripts to user-provided audit metadata (task purpose, outcome notes, and a satisfaction rating) and treating user purpose and user-perceived outcomes as first-class evaluative signals rather than optional feedback. This design allows for the analysis of users' own goals and judgments of success in their LLM interactions, producing evidence that is structurally absent from transcript-only collection and from existing in-interface feedback mechanisms alike. Our system comprises a browser extension (Figure \ref{fig:extension}), Django backend, and a deployment template with configurable evaluation report fields and instructions for self-hosted or cloud-hosted instances, all publicly available to support replication and adaptation by other communities.\footnote{\scriptsize{\url{https://github.com/victorojewale/monitrllm-browser-extension}}}

Second, we demonstrate MonitrLLM's viability through a two-week feasibility pilot with 26 college students interacting with ChatGPT \citep{openaichatgpt}, analyzing 206 audit reports in which participants contributed interaction data and outcome assessments. The pilot confirms that the infrastructure is deployable, that community-contributed metadata produces findings not recoverable from transcripts alone, and that some of those findings have direct implications for how LLM failure should be understood and measured.

\section{Background and Related Work}
\label{sec:related}

This section covers three bodies of literature that together motivate MonitrLLM's design: (1) how LLMs are currently evaluated through benchmarks and adversarial testing, (2) how human feedback and large-scale conversation data have been used to study model behavior in the wild, and (3) how communities and end users have been positioned as auditors of algorithmic systems

\subsection{Benchmarks and Red Teaming for LLM Evaluation}
A common approach to evaluating LLMs is benchmark-driven measurement \citep{chang2023surveyevaluationlargelanguage, hendrycks2020measuring, dac2023okapi,goyal2022flores}. Holistic benchmark suites such as HELM broaden evaluation beyond narrow task accuracy by measuring multiple desiderata, including robustness and fairness-related metrics, across many scenarios \citep{liang2023holisticevaluationlanguagemodels}. These efforts have made model comparison more systematic, but they still operationalize evaluation through predefined tasks and controlled prompting regimes. As a result, they only partially capture the open-ended, goal-driven work that characterizes everyday LLM use, where users bring idiosyncratic constraints, negotiate acceptable formats, and decide whether an output is usable for a particular context.

Red teaming complements benchmarks by probing model behavior under stress or adversarial conditions \citep{purpura-etal-2025-building}. For example, \citet{perez2022redteaminglanguagemodels} use language models to generate adversarial prompts that elicit undesirable behavior and expand coverage beyond what benchmark prompts typically include. However, red teaming remains largely expert-driven and is shaped by the hypotheses auditors choose to test. Even when red teaming surfaces important failure modes, it does not necessarily reveal how those failures manifest in routine workflows or how users adapt through iterative interaction, including repair strategies that turn partially correct outputs into usable results~\citep{10.5555/3716662.3716698}.

\subsection{Human Feedback, Conversation Collection and Data Donation}

A parallel line of work treats human feedback as a signal for model improvement rather than only as a measurement tool. Reinforcement learning from human feedback (RLHF) uses pairwise preference judgments to fine-tune model behavior toward outputs that human raters prefer \citep{10.5555/3600270.3602281,ziegler2020finetuninglanguagemodelshuman}. The preference signal in RLHF is efficient and scalable, but it captures whether a rater preferred one response over another in a comparison task, not whether a response fulfilled the user's underlying goal in a situated context. A response can be preferred by a rater without being usable by the person who actually submitted the query; we designed MonitrLLM to surface these failures that hide in this gap between rater preference and user-defined task success.

A second approach emphasizes collecting naturalistic conversations as a substrate for analysis, benchmarking, and alignment research. Datasets such as OpenAssistant and WildChat provide large corpora of human--LLM interactions that support research on deployment behavior \citep{köpf2023openassistantconversationsdemocratizing, zhao2024wildchat1mchatgptinteraction}, and LMSYS-Chat-1M \citep{zheng2024lmsyschat1mlargescalerealworldllm} extends this direction with one million real-world conversations spanning more than 25 LLMs, enabling large-scale analysis of prompting patterns and model behavior across a diverse user base. ShareLM further provides a browser plugin that enables users to voluntarily contribute chat transcripts with explicit review and sharing controls \citep{don-yehiya-etal-2025-sharelm}. These datasets and tools are valuable because they capture naturalistic prompting outside lab settings and at a scale larger than any controlled study.

At the same time, large-scale collection efforts optimize for breadth and general reuse, treating conversations as generic data points rather than as records of a particular community's interaction with a system it depends on. Community governance, community-defined success criteria, and context-specific questions about whether a model is serving a particular group well are largely absent from both the collection design and the resulting analyses. We argue that community-centered evaluation requires not only transcript collection, but also structured documentation of user intent and situated outcomes so that evaluation is anchored in what a user was trying to do and how they judged whether an interaction succeeded.

\subsection{End-User and Community-Led Audits}
Work on algorithm auditing grounds evaluation in questions of accountability and evidence rather than measurement alone. In practice this means querying systems to infer behavior from observed outputs when internal access is limited; later surveys formalizes this as an “outside-in” approach and show how audit goals, threat models, and evidence standards shape what audits can conclude \citep{Sandvig2014AuditingA,10.1561/1100000083}. Complementing external audits, \citet{raji2020closingaiaccountabilitygap} argue for end-to-end internal auditing processes that integrate documentation and value-based checks throughout system development and deployment.

In human-computer interaction research, a key shift in auditing research has been toward participatory and community-led auditing, motivated by the observation that impacted users often hold domain knowledge about what harms look like and which cases matter. \citet{10.1145/3555625} introduce end-user audits and show that non-technical community members can lead system-scale investigations when provided with scaffolds for hypothesis generation, evidence collection, and communication. This reframes auditing as a community capability rather than just as an expert service. 

MonitrLLM builds on these traditions by combining conversation logs with community-centered evaluative metadata. Like most data donation platforms, it captures naturalistic interactions, but it is designed around community deployment \emph{and treats user purpose and user-perceived outcomes as first-class evaluative signals rather than optional feedback}. Like end-user auditing systems, it aims to broaden what is visible by centering community questions and enabling stratified and thematic analysis grounded in a community's own accounts of success and failure.

\section{MonitrLLM: System Design}

MonitrLLM is a browser-based infrastructure for community-centered LLM evaluation consisting of a browser extension, a backend API, and a deployment template. Its status as a main contribution in this paper reflects our intention to present the infrastructure itself as a reusable artifact, not merely as the means by which our pilot data were collected.

\subsection{Design Considerations}
Community-centered LLM evaluation places specific demands on infrastructure that shaped the design of MonitrLLM. At minimum, such evaluation should be anchored in user goals and user-defined success. MonitrLLM operationalizes this by linking full conversation transcripts to a user's stated purpose and their judgment of whether the interaction succeeded. In practice, this meant capturing interaction text alongside ratings and outcome comments as a linked record.

Because community-centered evaluation depends on voluntary contribution, the system also needed to minimize submission burden. Data donation schemes are participation-sensitive and when contribution requires significant effort or workflow disruption, participation concentrates among users with particular motivations with the resulting data reflecting that bias
\citep{don-yehiya-etal-2025-sharelm}. Integrating the submission form directly into participants' existing LLM workflow, rather than requiring a separate platform, was the primary design response to this constraint.
 
A third requirement concerned ongoing participant control. Submitting interaction data raises consent questions that persist beyond initial enrollment \citep{10.1145/3551624.3555285}, so the system needed to make deletion and review accessible throughout participation.
 
Finally, because evaluation priorities differ across communities, the system needed to be reconfigurable, allowing different groups to tailor the tool's use to their purposes while using the underlying transcript-linkage infrastructure. The evaluation report form fields and target platform are therefore designed as independent configuration points.

\subsection{System Architecture and User Workflow}
 
We chose the web browser as the platform for MonitrLLM for the same reasons that motivate browser-based audit tools more broadly \citep{10.1145/3613904.3642473,lam2023intervenr}. Browsers sit at the intersection of a user's digital activity, require no additional software installation beyond the extension itself, and allow lightweight integration alongside the platform being evaluated without modifying it.
 
The MonitrLLM browser extension runs alongside the ChatGPT web interface and activates when a user clicks on its toolbar icon to submit an interaction. Each submission consists of five fields: a free-text interaction description, the user's task purpose, an outcome assessment, the ChatGPT share link for the conversation (validated by regular expression before acceptance), and a 1-to-5 satisfaction rating. The core design choice is to treat this metadata as \textit{evaluative signal}, with the transcript providing evidence of the interaction trajectory while the metadata captures the user's situated intent and their assessment of success, information that is otherwise absent from transcript-only collection.
 
On the backend, a Django API receives the submitted form. We then retrieve the full conversation transcript from the ChatGPT share link, a URL the user generates within ChatGPT to make a conversation's full history accessible outside the interface. The transcript and user-provided metadata are stored as a linked evaluation record. User identity is managed via a UUID generated on first install and stored locally in the browser's extension storage; this avoids account registration entirely, reducing participation friction and limiting the collection of personally identifying information. Participants can view and delete any of their submissions at any time through the extension's options page, preserving opt-out control after submission while being stored with anonymized identifiers and timestamps. The transcript-to-metadata linkage enables researchers deploying the system in their own use case to analyze failure patterns and interaction trends across task types, domains, and outcome ratings in ways that transcript-only corpora do not support.

\subsection{Deployment Template}
 
The browser extension and Django backend are released together as an open-source deployment template intended to lower the barrier to community-centered auditing in settings beyond the current pilot. The release includes the full extension source code, the backend API for receiving, storing, and deleting audit reports, and a deployment guide covering environment configuration and database setup for both self-hosted and cloud-hosted configurations.
 
Communities wishing to adapt MonitrLLM can reconfigure it along two axes without modifying the transcript-linkage core. The audit form fields (the purpose taxonomy, the outcome prompt, and the satisfaction scale) can be adjusted in the Django model and extension popup to reflect the community's own evaluation questions. The target platform can be changed by updating the URL validation regex, which currently accepts ChatGPT share links; supporting a different LLM interface requires only this change.

\section{Pilot Study Context and Methods}

To demonstrate MonitrLLM's viability, we conducted a two-week feasibility pilot with 26 college students, analyzing 194 audit reports covering ChatGPT interactions across academic and everyday tasks.

\subsection{Participants and Data Collection}

College students represent LLM user population whose tasks are diverse and whose outcomes are consequential enough (assignments, research, career work) to make their own judgments meaningful evaluation signals. Twenty-six participants were recruited through course announcements at a university in the Northeastern U.S., and compensated with participating in a raffle to get one of six \$50 gift cards, and the study was reviewed by the institution's IRB. Participants installed the MonitrLLM extension and instructed to use ChatGPT as they would normally, including academic and personal activities. They  were asked to submit reports for interactions they considered worth reflecting on, including both positive and negative experiences, and were encouraged to log at least one interaction each day over the study period without a fixed daily or total quota. We emphasized that sensitive or private conversations should not be submitted. The extension collects no data outside of interactions participants explicitly choose to submit, and participants were informed that their interactions would be anonymized prior to analysis. 

Across the study period, participants submitted 206 evaluation reports. We excluded conversations where the transcript was unavailable because the share link was deactivated after submission. After exclusions, 194 conversations from 25 participants form the working dataset.

A small subset of conversations (n=9) occurred partially or fully in Chinese. For these, we produced English versions via automated translation and validated each with a native speaker by cross-referencing the original transcript. Coders used the validated English translation as the reference.

\subsection{Analytical Approach and Coding Scheme}

We follow a sequential mixed-methods design~\cite{creswell2018choosingmixedmethods} in which we first familiarized ourselves with the dataset of collected audit reports, iteratively developed a codebook of conversation types and failure modes, and used it to systematically code all conversations before conducting quantitative analysis. Conversation excerpts are used to contextualize and interpret findings. 

Each conversation was coded along four dimensions chosen to capture the aspects of student LLM use most relevant to our evaluation questions: request type (capturing the nature of the request), topic/domain (the contextual purpose of the interaction), the subject matter involved, and the type of failure observed in the interaction. These dimensions reflect our priorities for this pilot; other communities deploying MonitrLLM could configure different dimensions depending on their own evaluation questions. Full definitions are included in Appendix~A. Two of the authors coded all conversations independently before adjudication.

\textit{Request type} captures the nature of the user's primary ask. We developed nine categories inductively through close reading of the evaluation reports: information and explanation, coding and debugging, writing and editing, mathematical reasoning, planning and advice, brainstorming, summarization, career support, and transcription. For example, a student asking the model to explain what an attenuated strain means in a biology paper would be coded as information and explanation, while a student asking for help debugging a loop would be coded as coding and debugging. Definitions for all request types are included in Table~\ref{tab:req_type} in the Appendix.

\textit{Topic domain} captures the contextual purpose of the interaction. We identified seven categories including academic coursework, academic research, personal and everyday use, career and job contexts, health and well-being, creative hobby, and unclear context. A student using the model to work through a homework problem would be coded as academic coursework; a student asking for dinner recommendations would be coded as personal and everyday use. Additional signals and examples are detailed in Appendix Table~\ref{tab:topic_domain}.

\textit{Subject matter} codes the content area of the interaction. We developed sixteen categories, including technical computing, mathematics and statistics, social science, natural science, humanities, and others. This dimension captures what the conversation is about, independent of the request type or domain. All subject matter categories and their corresponding descriptions are listed in Table~\ref{tab:subject-matter} in the Appendix.

\textit{Failure type} codes the primary failure mechanism observed in an interaction. We developed six categories: misinterpretation and reframing, interaction friction and non-convergence, dissatisfaction unspecified, factual error or hallucination, refusal or capability limit, and missing grounding or citation; conversations with no identifiable failure were coded \textit{none\_observed}. 
For example, a conversation where the model repeatedly answered a different question than the user asked would be coded as misinterpretation and reframing. We coded a conversation as a failure only when the participant's outcome note or satisfaction rating signaled that their goal was not met --- we do not infer failure from the transcript alone. For full definitions of all failure types, see Table~\ref{tab:failure-types} in the Appendix.

\subsection{Coding Procedure and Adjudication}
The first two authors coded all conversations independently followed by a fixed evidence review order to ensure labels reflected participants' situated success criteria. We first reviewed each participant's outcome notes and purpose statement to establish their intended task and success criteria, then read the full transcript in light of those signals before assigning labels. For ambiguous cases, we recorded a one-sentence evidence note identifying the specific transcript span or outcome note that motivated the label.

Disagreements were resolved through structured adjudication. After independently coding, we met to compare codes and calculate percentage agreement. We discussed retained disagreements with reference to specific transcript evidence and outcome notes until reaching full agreement. No case required more than two rounds to resolve.

\section{Pilot Findings: What the Infrastructure Surfaces}
\label{sec:findings}

The findings that follow draw on 194 evaluation reports collected over two weeks from 25 college students. They should be read as demonstrations of what the infrastructure makes analytically visible rather than as generalizable claims about student LLM use; each depends structurally on at least two of MonitrLLM's three linked data sources (the conversation transcript, the user-reported task purpose, and the outcome assessment)

\subsection{Student-LLM Interactions as Evaluation Context}
\begin{figure}[t]
  \centering
  \includegraphics[width=0.49\textwidth]{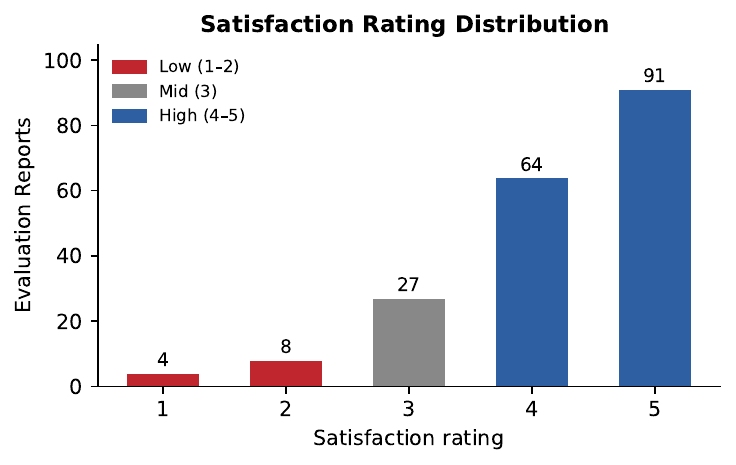}
  \caption{Frequency distribution of user satisfaction ratings (1 = lowest, 5 = highest) across the evaluation reports.}
  \label{fig:ratings}
\end{figure}

Overall satisfaction across 194 rated conversations was high (see Figure \ref{fig:ratings}), with a mean rating of 4.19 ($SD = 0.96$, median = 4). 79.9\% of conversations were rated 4 or 5, and only 6.2\% were rated 1 or 2. Successful interactions were often described in direct terms. One participant who used the model for a research explanation noted that the response was ``helpful and even helped with troubleshooting'' and that they ``found it did not hallucinate.'' A student who worked through a coding problem reported ``It fixed my problem! It was a small syntax one.'' These represent the modal experience in the sample; the analysis below begins with the general composition of the dataset before turning to findings that require all three linked sources.

\paragraph{Request type.}
Table~\ref{tab:reqtype} shows the distribution of request types across 194 coded conversations (see Table~\ref{tab:req_type} in Appendix for request type definitions). Information and explanation requests were the largest category at 40.2\% ($n=78$), reflecting students seeking conceptual clarification, background, and
explanations across a range of subject areas. Writing and editing (13.4\%, $n=26$) and planning and advice (13.4\%, $n=26$) tied for second, followed by coding and debugging (12.9\%, $n=25$) and mathematical reasoning (8.2\%, $n=16$). The  distribution reflects the breadth of everyday student work.

\paragraph{Topic domain and subject matter.}
Academic contexts dominated the topic domain and subject matter. Academic coursework comprised 31.8\% of conversations ($n=62$) and academic research 19.5\% ($n=38$), together accounting for just over half of all interactions. Personal and everyday use accounted for 30.8\% ($n=60$), with career and workplace (8.2\%), unclear context (8.2\%), and creative or health contexts making up the remainder. The largest subject matter categories were technical computing (23.7\%, $n=46$), general and other (18.6\%, $n=36$), mathematics and statistics (11.3\%, $n=22$), and social science (9.8\%, $n=19$). Definitions and descriptions for topic/domain and subject matters are included in Tables~\ref{tab:topic_domain} and \ref{tab:subject-matter}, respectively.

\paragraph{Task profiles by domain.}
The cross-tabulation of request type by topic domain reveals distinct task profiles per context. Academic research conversations were weighted toward information and explanation (44.7\%) and coding and debugging (21.1\%). Academic coursework showed a
more even spread across information seeking (22.6\%), writing and editing (21.0\%), and mathematics (21.0\%). Personal and everyday conversations were dominated by information seeking (60.0\%) and planning (25.0\%). Career and workplace conversations were the only domain with substantial writing and editing (31.2\%) and career support (18.8\%) shares, consistent with the domain's focus on professional positioning and communication.

\begin{table*}[t]
\small
\centering
\begin{tabular}{lrrrr}
\toprule
Request type & $n$ & Share & Mean rating & Fail rate \\
\midrule
information\_explanation  & 78 & 40.2\% & 4.13 & 26.9\% \\
writing\_editing          & 26 & 13.4\% & 4.15 & 26.9\% \\
planning\_advice          & 26 & 13.4\% & 4.00 & 11.5\% \\
coding\_debugging         & 25 & 12.9\% & 4.16 & 32.0\% \\
math\_reasoning           & 16 &  8.2\% & 4.69 & 11.8\% \\
brainstorming             & 13 &  6.7\% & 4.15 & 15.4\% \\
summarization             &  6 &  3.1\% & 4.50 & 33.3\% \\
career\_support $\dagger$ &  3 &  1.5\% & 4.67 &  0.0\% \\
transcription $\dagger$   &  1 &  0.5\% & 4.00 &  0.0\% \\
\bottomrule
\end{tabular}
\caption{Request type distribution across 194 coded conversations. Categories with $n < 5$ are marked $\dagger$ and interpreted with caution.}
\label{tab:reqtype}
\end{table*}

\subsection{How LLM Use Plays Out}
We now examine how interactions unfolded in practice and what that reveals about failure, drawing on transcript structure and participant reported outcomes together.

\paragraph{Iterative prompting is the norm.}
61.5\% of conversations included at least one follow-up prompt beyond the initial exchange. This varied substantially by task type. Coding and debugging showed the highest follow-up rate in this sample (80.0\%) and the longest mean turns per conversation (5.32), followed by writing and editing (69.2\% follow-up rate, mean 5.88 turns). Planning and advice tasks had a similarly high follow-up rate (69.2\%) but much shorter conversations (mean 2.81 turns). Brainstorming showed the lowest observed follow-up rate at 30.8\%. By topic domain, academic research (71.1\%) and academic coursework (69.4\%) showed the highest follow-up rates.

\paragraph{Follow-up suggests failure.}
Conversations with follow-up prompting had a participant-reported failure rate of 30.0\%, compared to 12.0\% for single-turn conversations. A possible assumption that more prompting reflects deeper engagement inverts in this data; follow-up conversations failed at 2.5 times the rate of single turn, marking extended interaction as a signal of difficulty rather than productive use. Among failed conversations specifically, the cost is visible in turn counts: failed conversations involved on average 6.11 user turns ($SD = 7.93$, median = 4) compared to 3.12 for successful ones ($SD = 5.09$, median = 2), a pattern that held across analyses capping turn counts at 10, 15, and 20 to rule out outlier influence.

When a user receives a misinterpreted or inadequate response, they face a choice between abandoning the task, switching tools, or investing further turns to repair the interaction, a decision point that one participant described plainly after a long coding session:
``I felt like it was too much work to continue troubleshooting so I stopped.'' Another wrote that they ``got frustrated and then just resorted to simple tasks that I knew chatgpt could do well.''
 
The gap between turn count and actual experience is sharpest in a case from this pilot where a participant engaged across 16 follow-up exchanges, rated their conversation a 4 out of 5, and described the responses as ``helpful, but wasn't accurate about 10\% of the time,'' noting inaccurate citations and facts. Nothing in the transcript flagged this as a failure: the conversation appeared productive, the rating was positive, and the failure was only visible in the outcome note. Transcript evidence can be structurally consistent with success while the user is quietly managing a model that is unreliable in ways they can detect but the log cannot, and turn count is ambiguous at best and misleading at worst unless it is read against the user's own judgment of whether the interaction succeeded.

\begin{table*}[t]
\small
\centering
\begin{tabular}{lrrr}
\toprule
Failure type & $n$ & \% of failures & Mean rating \\
\midrule
None observed (baseline)        & 149 & ---    & 4.54 \\
\midrule
Misinterpretation and reframing &  15 & 33.3\% & 2.80 \\
Dissatisfaction unspecified    &  11 & 24.4\% & 3.18 \\
Interaction friction / nonconv. &  10 & 22.2\% & 2.60 \\
Factual error or hallucination  &   7 & 15.6\% & 3.57 \\
Refusal or capability limit     &   1 &  2.2\% & 4.00 \\
Missing grounding or citation   &   1 &  2.2\% & 4.00 \\
\midrule
\textbf{Any failure (total)}    & \textbf{45} & 100\% & \textbf{3.02} \\
\bottomrule
\end{tabular}
\caption{Failure taxonomy across 194 coded conversations. Conversations with no observed failure ($n=149$, mean rating 4.54) shown as baseline. Overall failure rate 23.1\% (45 of 194). Satisfaction gap $\Delta = 1.52$ points.}
\label{tab:taxonomy}
\end{table*}

\subsection{A Transcript-Grounded Failure Taxonomy}

The overall observed failure rate was 23.1\% (45 of 194 conversations). Failure-free conversations had a mean satisfaction rating of 4.54, compared to 3.02 for conversations with any observed failure, a gap of 1.52 rating points. Table \ref{tab:taxonomy} presents the full taxonomy with frequencies and mean satisfaction ratings. Definitions per type are included in Table~\ref{tab:failure-types} in the Appendix.

A methodological observation applies to all six failure categories. Most failures are only legible when the transcript is read alongside the user's outcome note. A response can appear plausible in isolation but be unusable in context because it misses the
user's actual goal, cannot be verified in the intended setting, or asserts information the user has no means to check. This is the core empirical argument for why community audit metadata changes what evaluation can see.

\textbf{Misinterpretation and reframing} was the most frequently observed failure type in this sample, appearing in 15 conversations (33.3\% of all failures, mean rating 2.80). The model answers a different question than intended, prompting users to restate constraints or redirect across multiple turns. A student who used the model for coding troubleshooting wrote
that it ``talked in circles and did not consider other possibilities,'' and that they ended up troubleshooting themselves before returning to ask further questions. Another participant, who had pasted a quoted passage for grammatical analysis, was ``slightly frustrated that it didn't understand that the prompt I gave it, with text in quotes and using bracketed ellipses to show omission, was clearly a quote.'' A writing task ended with the note that the model ``didn't really get what I wanted writing-wise but the feedback was still helpful,'' illustrating how partial success and misinterpretation can coexist within the same interaction.

\textbf{Dissatisfaction unspecified} was the second most frequently coded type, observed in 11 conversations (24.4\%, mean rating of 3.18). This is a residual category for cases where the user signals a negative outcome through a low rating or a negative note, but neither the transcript nor the notes provide sufficient evidence to assign a specific mechanism. One participant described a brainstorming response as ``really generic,'' adding that this was ``inevitable and rather preferable considering it doesn't actually know much about me.'' Another participant noted that their writing output ``still sounded very AI.'' These notes signal dissatisfaction without pointing to a detectable failure mode in the transcript, which is itself a finding about the limits of transcript-only evaluation. That these cases don't fall neatly into the other failure categories acknowledges that a negative outcome occurred even when neither the transcript or outcome notes can explain why. That is already a step beyond what transcript evidence alone could surface, where the same cases might be dismissed if no clear reason was provided.

\textbf{Interaction friction and nonconvergence} appeared in 10 conversations (22.2\%, mean rating 2.60), denoting exchanges where repeated re-prompts fail to yield a usable output. Signals include escalating specificity and outcome notes emphasizing time cost. One participant described a coding session in which the model ``would give me entirely new commands instead of just tweaking the old ones'' and eventually switched programming languages after the issue remained unresolved. Another, working on a data analysis task, wrote that the model ``made more complicated code than necessary'' and they ``ended up doing the coding myself.'' This failure type was concentrated in coding and debugging conversations (5 of 8 failures in that category) and in academic research (6 of 12 failures), consistent with the high mean turn counts observed in those domains.

\textbf{Factual errors and hallucinations} appeared in 7 conversations (15.6\%, mean rating 3.57). The relatively higher mean rating here, compared to misinterpretation and friction, likely reflects cases where users only partially detected the error.
One user described how the model was ``incorrect many times'' and did not give an accurate answer on a factual question. Another user flagged hallucinated BibTeX entries, noting that the model ``would hallucinate wrong source sometimes when given link'' but would fix the entry when presented with the correct title. These notes vary in how completely users detected the error, suggesting that 7 is a lower bound on this failure type.

\textbf{Refusal or capability limit} and \textbf{missing grounding or citation} each appeared once. The refusal case involved a user who had hoped for a solution within Google Docs but found the model unable to operate within those constraints. The missing grounding case appeared in an academic research context where a user who requested 25 reputable and updated sources found that ``only 4/25 sources found by deep research were useful.'' This single case almost certainly understates
the true rate of grounding failures, since users who accept unverifiable responses without noting the gap would not produce outcome notes that surface this pattern.

\subsection{Where Failure Concentrates}

\begin{table*}[t]
\small
\centering
\caption{Observed failure rates by request type (left) and topic domain (right). All comparisons are descriptive.}
\label{tab:failure}
\begin{tabular}{lrrr}
\toprule
Request type & $n$ & Fail rate & Mean rating \\
\midrule
summarization         &  6 & 33.3\% & 4.50 \\
coding\_debugging     & 25 & 32.0\% & 4.16 \\
writing\_editing      & 26 & 26.9\% & 4.15 \\
information\_expl.    & 78 & 26.9\% & 4.13 \\
brainstorming         & 13 & 15.4\% & 4.15 \\
math\_reasoning       & 16 & 11.8\% & 4.69 \\
planning\_advice      & 26 & 11.5\% & 4.00 \\
\bottomrule
\end{tabular}
\quad
\begin{tabular}{lrrr}
\toprule
Topic domain & $n$ & Fail rate & Mean rating \\
\midrule
academic\_research     & 38 & 31.6\% & 4.13 \\
academic\_coursework   & 62 & 29.0\% & 4.16 \\
\textit{Academic combined} & \textit{100} & \textit{30.0\%} & \textit{4.15} \\
personal\_everyday     & 60 & 21.7\% & 4.17 \\
unclear\_context       & 16 & 12.5\% & 4.25 \\
career\_job\_workplace & 16 &  0.0\% & 4.25 \\
creative\_hobby        &  2 &  0.0\% & 5.00 \\
\bottomrule
\end{tabular}
\end{table*}

\paragraph{Failure by request type.}
Table~\ref{tab:failure} (left panel) shows observed failure rates by request type for categories with $n \geq 5$. Coding and debugging showed the highest observed failure rate among categories with $n \geq 5$ in this sample (32.0\%), more than 2.7 times that of mathematical reasoning (11.8\%) and planning and advice (11.5\%). Information and explanation tasks
had an observed failure rate of 26.9\%, as did writing and editing. The interaction friction failure type was concentrated in coding conversations, consistent with the account that coding tasks are environment-specific and iterative in ways that math and planning tasks are typically not. A contrast visible in the outcome notes captures this directly. A student who used the model for a math problem wrote ``It was extremely helpful! I got my answer,'' while a student in a coding session wrote
that the model ``was not able to figure out the issue'' and they switched languages.

\paragraph{Failure by topic domain.}
Table~\ref{tab:failure} (right panel) shows failure rates by topic domain. Academic research (31.6\%, $n=38$) and academic coursework (29.0\%, $n=62$) showed the highest observed failure rates in this sample, with academic contexts combined at
30.0\% ($n=100$). Personal and everyday conversations showed a lower observed rate (21.7\%, $n=60$). Career and workplace conversations showed no observed failures in this sample ($n=16$), though the small category size means this should be interpreted cautiously.

This raises an important point with regards to the stakes of usage in different domains. Academic researchers rely on precision, needing verifiable claims, citable sources, and outputs that hold up under scrutiny. One research participant's note about section numbering illustrates the stakes: ``a few of the section numbers that it gave were based on an older edition, so the section numbering was inconsistent. This was not too difficult to work around, but notably wrong.'' That kind of error is subtle enough to go undetected without domain knowledge, but very consequential context. The aggregate satisfaction mean of 4.19 across all conversations conceals this heterogeneity.

\section{Discussion}

\paragraph{What community-centered evaluation makes visible.}
The central contribution of this paper is an infrastructure for community-centered LLM evaluation, and the pilot findings are best understood as a demonstration of what that infrastructure makes visible rather than as self-contained empirical claims. MonitrLLM preserves three things that existing evaluation mechanisms do not link together, namely the full interaction trajectory, the user's stated task purpose, and the user's own judgment of whether the interaction succeeded, and the key findings of this study are only recoverable when all three are read in combination. The 23.1\% failure rate is invisible in the aggregate satisfaction mean of 4.19 and only surfaces when outcome assessments are applied as a stratifying variable across rated conversations. The 2.5-fold difference in failure rates between follow-up and single-turn conversations only emerges when trajectory structure is read against user-reported success rather than treated as a proxy for engagement, and the 2-to-1 ratio in mean user turns between failed and successful interactions only becomes interpretable when turn count is anchored to outcome rather than counted as an absolute measure of interaction. The concentration of failure in academic contexts, at 30.0\% compared to 21.7\% for personal and everyday use, is also  visible when domain metadata is part of the evaluation record. What these findings share is that they do not require a better model or a new benchmark but only that the right evidence be preserved in the first place.

\paragraph{A mixed-methods analysis of student LLM use.}
The findings are the product of a sequential mixed-methods design \citep{creswell2018choosingmixedmethods} in which qualitative coding of student LLM use interaction transcripts and outcome notes precedes and structures quantitative analysis, and this design is not incidental but reflects a methodological commitment that is constitutive of the community-centered approach. Quantitative patterns establish prevalence and enable comparison across task types and domains, but they do not identify mechanisms, and the failure taxonomy in Table \ref{tab:taxonomy} is a qualitative product whose six categories emerged from careful reading of transcripts alongside outcome notes and whose meanings are carried by the illustrative excerpts in the findings section as much as by the frequency counts. The multi-turn burden finding illustrates the complementarity of the two modes most directly, since the quantitative pattern of a 2-to-1 turn ratio, stable across outlier-capping thresholds, identifies that something is systematically different about failed conversations, but it is the outcome notes that explain what that difference is: participants who stayed through long sessions were often working around a model that had misunderstood their task rather than elaborating a successful one, and that interpretation is only available because the outcome metadata accompanies the transcript. Together, the two sources support analyses that would not be possible from transcript-only corpora, from in-interface satisfaction ratings, or from benchmark evaluation taken alone.

\paragraph{MonitrLLM as a standalone and complementary evaluation tool.}
The infrastructure is designed to be useful in two distinct (but not mutually exclusive) deployment scenarios.

As a \textit{standalone evaluation tool}, MonitrLLM supports community-centered evaluation in settings where no other systematic assessment is in place. For communities that rely on LLMs for high-stakes work but have no mechanism for surfacing failures, the infrastructure provides a structured way to collect evidence, develop a community-grounded failure taxonomy, and track whether failure patterns change over time or across use contexts, and the pilot demonstrates that this is viable at small scale, since 26 participants over two weeks proved sufficient to produce actionable failure patterns across task types and domains.

As a \textit{complementary tool}, MonitrLLM adds a situated-use dimension to evaluation frameworks that measure model capabilities without observing how those capabilities translate to real tasks. Benchmark evaluation can establish what a model is capable of under controlled conditions, while MonitrLLM surfaces whether those capabilities serve users when the task is self-defined, the context is idiosyncratic, and success is judged by the user rather than a fixed correctness criterion. While reinforcement learning from human feedback \citep{10.5555/3600270.3602281} captures which response a rater prefers in a direct comparison, MonitrLLM can capture whether a response fulfilled the underlying goal of a user pursuing a real task across multiple turns. Red teaming identifies failures under adversarial or stress conditions, while MonitrLLM identifies failures in ordinary use, where users are not trying to break the model but simply trying to get their work done. The failure types surfaced by MonitrLLM, including misinterpretation and reframing, nonconvergence, and unverifiable citations, are not the failure modes that adversarial probing is designed to find, and the two methods therefore provide complementary coverage rather than redundant evidence.

\paragraph{Implications for evaluation practice.}
The findings in this study come from a single institution over two weeks, but the infrastructure argument they support is not bounded by those conditions. As LLMs become more embedded in consequential knowledge work, the gap between what benchmark and preference-based evaluation can detect and what users actually experience in context is likely to widen rather than narrow, since models improve on held-out tasks while the range of situated goals users bring to them continues to expand, and closing this gap requires investment in the evidence layer, meaning the mechanisms that determine what information is preserved when users interact with AI systems and on what terms that information can be queried. This is an institutional and organizational challenge as much as a technical one, since deploying MonitrLLM in a new community requires configuring a form and standing up a backend, but sustaining it requires that the community has standing to act on what the audit records reveal, that data ownership is clearly assigned, and that the evaluation questions the infrastructure answers are the community's own. The open-source release addresses the technical barrier, and we see the deployment template as a starting point for communities to develop the institutional arrangements that make community-centered auditing a durable practice rather than a one-time study.

\section{Limitations and Future Work}
The community this study draws from is a single student cohort, which may limit generalizability in the conventional sense, though in a practical sense the concern applies differently here than it would for a study making claims about student behavior in general. The taxonomy, coding procedures, and audit infrastructure are designed to be transferrable, and the pilot's contribution is a demonstration of what the infrastructure makes visible rather than a characterization of any particular community's LLM use. That said, the pilot is scoped as a feasibility study, and several design decisions create boundaries on the current findings that future deployments can address.

The most important boundary concerns submission self-selection as participants submitted reports for interactions they found worth reflecting on rather than a random sample of their activity, which means the observed failure rate reflects the failure rate among salient interactions rather than a population-level estimate. A natural extension would pair voluntary reporting with an optional random-sampling mode that captures a random fraction of conversations automatically, allowing failure rates across the two conditions to be compared directly.

The two-week window is well-suited to a feasibility study but does not reveal how interaction patterns or failure rates change as users accumulate experience with LLMs; a longitudinal deployment, accompanied by periodic community sessions in which participants review interim findings and update their evaluation questions, would address that gap.

A more fundamental tension concerns the design of the audit form itself. Community-centered evaluation, as we describe it, should be anchored in the community's own evaluation questions, yet the form fields, coding dimensions, and failure categories used in this pilot were developed by the research team rather than through a co-design process with participants. An extended community-centered deployment would involve the community members in defining what to measure before data collection begins, for instance through focus groups or participatory design sessions that surface the failure modes and success criteria that matter most to that group. Future deployments of MonitrLLM should treat the audit form as a site of participatory design rather than just a researcher-configured instrument.

Finally, the open-source release of MonitrLLM's infrastructure makes community deployment technically accessible but leaves the governance questions open. Genuine community-centered evaluation requires that communities set their own evaluation questions, interpret findings, and hold institutional standing to act on them, and none of that follows from infrastructure alone. Future deployments are an opportunity to develop and document governance models (specifying data ownership, access controls, and the obligations deploying institutions hold toward contributors) that can accompany the technical infrastructure and make community-centered auditing a viable practice rather than a design aspiration.

Future extensions of MonitrLLM could broaden coverage in two directions. Adapting the infrastructure to support platforms beyond ChatGPT would allow community auditors to ask whether a given failure pattern is a property of the model, the interface, or both, a question the current single-platform design cannot answer. The audit form itself could also be extended with mechanisms for users to add clarification when they report low satisfaction, which would help reduce the \textit{dissatisfaction unspecified} residual category and give future coders more to work with.

\section{Conclusion}

We introduced MonitrLLM, infrastructure for community-centered LLM evaluation that pairs conversation transcripts with user-reported task intent and outcome assessments, and we reported on a two-week feasibility pilot in which 26 college students submitted 194 audit reports covering ChatGPT interactions across
a wide range of academic and everyday tasks.

The central argument of this paper is infrastructural, grounded in the observation that evaluation can surface only what the surrounding system preserves as evidence, and that the feedback mechanisms built into current LLM interfaces record whether users found a response satisfying without capturing what they were trying to accomplish, whether the output was usable for their specific context, or how much interaction effort the exchange required. MonitrLLM treats user intent and user-judged outcomes as primary evaluative signals rather than optional metadata, and linking those signals to full transcripts changes what becomes measurable. A 23.1\% failure rate invisible in an aggregate satisfaction mean of 4.19 becomes visible once outcome assessments are paired with transcript trajectories. The 2.5-fold difference in failure rates between follow-up and single-turn conversations, and the 2-to-1 ratio in mean user turns between failed and successful interactions, only emerge when trajectory length is read against user-reported success rather than treated as a proxy for engagement. The concentration of failure in academic contexts, where verifiability requirements are highest, is only visible when domain metadata is part of the evaluation record. These findings establish that community audit metadata changes what evaluation can see, and they do so in a setting where the stakes are direct, since students using LLMs for coursework and research face real costs when models misinterpret their requests, generate unverifiable citations, or fail to converge after repeated re-prompting, and aggregate satisfaction scores conceal those costs entirely.

As LLMs become more deeply embedded in consequential knowledge work, evaluation infrastructure that captures who uses these systems, for what purpose, with what outcome, and at what cost in interaction effort becomes a necessary complement to benchmark evaluation. MonitrLLM's open-source release is one concrete step toward that infrastructure, and we see it as a foundation that other communities can build on to ask their own evaluation questions on their own terms.

\section*{Acknowledgments}
We thank the students who participated in this pilot for contributing their conversations and outcome assessments. We are also grateful to members of the Center for Tech Responsibility and PennHCI who provided feedback on earlier designs of the Browser Extension. This work was supported in part by the MacArthur Foundation and the Heising-Simons Foundation.

\bibliography{aaai2026}

\clearpage
\onecolumn

\appendix

\section{Appendix}
\label{appx:defs}

\subsection{Request types and definitions}
The request types detailed in Table~\ref{tab:req_type} were used to code the kinds of requests students engaged in when interacting with ChatGPT. They encompass the overall intention the users had during the conversation including seeking information or explanation on a topic, assistance coding and debugging, and support for tasks such as writing or mathematical problem solving. Initial request types were derived from Wildchat transcripts~\cite{zhao2024wildchat1mchatgptinteraction} and iteratively refined through analysis of MonitrLLM user conversation transcripts as additional request types surfaced, until thematic saturation was reached.

\renewcommand{\arraystretch}{1.5}
\begin{table*}[h]
\centering
\small
\begin{tabularx}{\textwidth}{@{}lX@{}}
\toprule
\textbf{Request type} & \textbf{Definition} \\ \midrule
information\_explanation & User primarily asks for an explanation, definition, background, or conceptual clarification. Output is meant to teach or explain. \\
summarization & User asks to condense, paraphrase, extract key points, or produce a structured summary of provided or referenced material (text, notes, article, transcript). \\
coding\_debugging & User asks for programming help for writing code, debugging errors, explaining code behavior, implementing an algorithm, fixing environment issues. \\
writing\_editing & User asks to draft, rewrite, polish, rephrase, adjust tone, or structure writing (emails, essays, statements, slides text). Includes "make this shorter", "rewrite in my tone", formatting requests. \\
math\_reasoning & User asks for mathematical problem solving or step-by-step reasoning (calculus, algebra, probability), including checking a solution. \\
planning\_advice & User asks for plans, recommendations, schedules, checklists, decision support, or next steps (study plan, itinerary, what to do). \\
brainstorming & User asks for idea generation, creative directions, alternatives, topic suggestions, naming, outlines as ideation rather than drafting. \\
career\_support & User asks for job or professional support: resume, cover letter, interview prep, career decisions, application positioning, workplace communication. \\
translation\_language & User asks for translation, wording in another language, language learning, phrasing, or multilingual rewriting. \\
transcription & User asks for assistance transcribing text in an image. \\ \bottomrule
\end{tabularx}
\caption{Request types and definitions used to code students' interactions with ChatGPT.}
\label{tab:req_type}
\end{table*}

\subsection{Topic/domain (context of use)}
The topic and domain of a conversation provided an additional layer of context for a particular user request. These topics were also iteratively refined after reviewing conversation transcripts. For example, a user could be seeking advice for specfic health symptoms, which would be a planning/advice request directly under the topic of health and wellbeing, or `health\_wellbeing', in our codebook. Refer to Table~\ref{tab:topic_domain} for a complete list of these topics and domains and their definitions.

\renewcommand{\arraystretch}{1.2}
\begin{table*}[h]
\centering
\small
\begin{tabularx}{\textwidth}{@{}lX@{}}
\toprule
\textbf{Topic/domain} & \textbf{Definition} \\ \midrule
academic\_coursework & Work intended for a class, assignment, exam prep, problem set, lab, or course project. \ Signals like \ ``homework'', ``assignment'', ``class'', ``course'', ``problem set'', ``exam'', ``professor'', ``due''. \\
academic\_research & Work intended for research, including literature review, paper writing, methods, experiments, or thesis/dissertation work. Signals: ``paper'', ``literature review'', ``related work'', ``methods'', ``study design'', ``thesis'', ``dissertation'', ``IRB'', ``analysis for research''. \\
career\_job\_search/workplace & Work intended for applying for roles or career positioning. Signals: ``resume'', ``CV'', ``cover letter'', `` interview'', ``application'', ``personal statement'', ``portfolio''. \\
personal\_everyday & Personal life coordination and low-stakes practical tasks like recipes, etiquette, personal scheduling, casual planning. Signals: recipes, travel planning, scheduling, etiquette messages, errands, general curiosity not tied to school or work. \\
health\_wellbeing & Health and wellbeing decisions for self or close others.  Signals: symptoms, diagnosis, medication, therapy, mental health, fitness advice sought as personal guidance. \\
creative\_hobby & Creative projects or hobbies done for enjoyment or personal creative output.  Signals: story, poem, design idea, branding for fun, creative writing prompts. \\
unclear\_context & Not enough information to infer intended setting, even using purpose, outcome notes, and transcript. \\ \bottomrule
\end{tabularx}
\caption{Topic/domain and definitions used to classify the context in which students interacted with ChatGPT.}
\label{tab:topic_domain}
\end{table*}

\subsection{Subject matter}
The subject matter of a user interaction pertains to the specific subject within a broader topic or domain, as described in Table~\ref{tab:subject-matter}. This level of coding was particularly useful for niche interactions that would have otherwise be obscured within broader request type and topic/domain categoies. As an illustrative example, many users in our pilot study used ChatGPT for information requests related to academic research that include subjects such as algorithm theory under `computer\_science', which is different than researching software tools, and would fall under `technical\_computing' instead.

\renewcommand{\arraystretch}{1.5}
\begin{table*}[h]
\centering
\small
\begin{tabularx}{\textwidth}{@{}lX@{}}
\toprule
\textbf{Subject matter} & \textbf{Description} \\ \midrule
technical\_computing & Programming, debugging, software tools, data analysis pipelines, and machine learning tooling. \\
math\_statistics & Mathematics and statistics content independent of coding. \\
natural\_science & Biology, chemistry, physics, materials science, and related scientific domains. \\
social\_science & Economics, sociology, anthropology, psychology, education theory, and related fields. \\
humanities & History, philosophy, literature, religion, classics, and related humanities disciplines. \\
govt\_law\_policy & Law, regulation, compliance, governance frameworks, and government agencies. \\
politics\_elections & Political parties, elections, ideology, social movements, partisan debate, and international relations. \\
personal\_finance & Budgeting, investing, taxes, retirement, and other personal financial topics. \\
health\_wellbeing & Medical, health, fitness, and wellbeing content. \\
personal\_relationships & Family, friendships, romance, interpersonal dynamics, and self-perception. \\
creative\_media\_design & Creative writing, art, design, storytelling, film, and related media production. \\
sports & Sports, athletics, teams, and related topics. \\
food\_cooking & Cooking, recipes, food preparation, dining, and culinary topics. \\
computer\_science & Computer science theory, algorithms, systems, and additional computing concepts. \\
travel & Travel planning, destinations, transportation, lodging, and tourism. \\
general\_other & Mixed, uncategorizable, or unclear subject matter. \\ \bottomrule
\end{tabularx}
\caption{Subject matter categories and corresponding descriptions used to code the primary content of students' ChatGPT interactions.}
\label{tab:subject-matter}
\end{table*}

\subsection{Failure types}
Coding for failure types was a critical component of this framework. Failure types allowed us to encode the kinds of failures users described from their chatbot interactions that were not always apparent from transcript review alone, or that provided additional context beyond the transcripts themselves. Our failure type taxonomy and definitions are detailed in Table~\ref{tab:failure-types}.

\renewcommand{\arraystretch}{1.5}
\begin{table*}[h]
\centering
\small
\begin{tabularx}{\textwidth}{@{}lX@{}}
\toprule
\textbf{Failure type} & \textbf{Definition} \\ 
\midrule
none\_observed & No clear breakdown surfaced in transcript or outcome notes. The user either succeeded or did not signal a meaningful issue. \\
factual\_error\_or\_hallucination & Model provides incorrect claims or fabricated specifics. Evidence can be user correction, contradiction with provided ground truth, or outcome notes stating it was wrong. \\
missing\_grounding\_or\_citation & The user explicitly requests sources, citations, quotes, links, or verifiable evidence, and the response fails to provide adequate grounding for the user's purpose. This is about verifiability, not necessarily correctness. \\
misinterpretation\_and\_reframing & Model answers a different question than intended or adopts an unhelpful framing; user has to restate, add constraints, or redirect. \\
interaction\_friction\_nonconvergence & Multi-turn interaction fails to reach a usable result despite repeated attempts, escalating constraints, or repeated re-prompts. Often paired with outcome notes about time cost or ``still not what I need''. \\
refusal\_or\_capability\_limit & Model refuses, declines, or signals inability (policy, missing access, feature limitation). User responds by narrowing, rephrasing, or switching tasks. \\
dissatisfaction\_unspecified & User signals dissatisfaction (rating low or outcome note negative) but transcript and notes do not provide enough evidence to assign a specific failure type. This is a ``we cannot tell why'' bucket, not a failure mechanism. \\
\bottomrule
\end{tabularx}
\caption{Failure types taxonomy used in our analysis, including operational definitions for interaction breakdowns, grounding failures, factual errors, refusals, and unresolved user dissatisfaction.}
\label{tab:failure-types}
\end{table*}

\end{document}